\documentclass[a4paper,11pt]{article}
\pdfoutput=1 

\usepackage[colorinlistoftodos,prependcaption,textsize=tiny]{todonotes}
\usepackage{jheppub} 
\usepackage[T1]{fontenc} 
\usepackage{amsmath}
\usepackage[utf8]{inputenc}

\usepackage{xcolor}

\usepackage{pythontex} 
\usepackage{listings}
\usepackage{amssymb} 
\usepackage{graphicx}
\usepackage{amsmath}
\graphicspath{ {./images/} }

\newcommand{\roughly}[1]{\mathrel{\raise.3ex\hbox{$#1$\kern-0.85em
\lower1ex\hbox{$\sim$}}}}

\newcommand{\be}{\begin{equation}}
\newcommand{\bee}{\begin{equation}}
\newcommand{\ee}{\end{equation}}
\newcommand{\beea}{\begin{eqnarray}}
\newcommand{\eea}{\end{eqnarray}}
\newcommand{\bea}{\begin{eqnarray}}

\def\LL{\mathcal{L}}

\title{Extremal learning: extremizing the output of a neural network in
regression problems}

\author[a]{Zakaria Patel}
\author[b]{and Markus Rummel}
\affiliation[a]{Engineering Physics, McMaster University, Hamilton, ON, Canada,
L8S 4M1}
\affiliation[b]{AI Endurance Inc, Hamilton, ON, Canada, L8P 0A1}

\emailAdd{patelz6@mcmaster.ca}
\emailAdd{markus@aiendurance.com}

\abstract{
Neural networks allow us to model complex relationships between variables. We
show how to efficiently find extrema of a trained neural network in regression
problems. Finding the extremizing input of an approximated model is formulated
as the training of an additional neural network with a loss function that
minimizes when the extremizing input is achieved. We further show how to
incorporate additional constraints on the input vector such as limiting the
extrapolation of the extremizing input vector from the original training
data set. An instructional example of this approach using TensorFlow is included.
}

\makeatletter
\gdef\@fpheader{}
\makeatother
\begin{document}

\maketitle
\flushbottom
\section{Introduction}
\label{sec:intro}
Neural networks (NNs)~\cite{McCulloch:1943} allow us to model complex
relationships between certain input $(x_n)$ and output data sets $(y_n)$ when the
underlying functional dependence is unknown. Advances in hardware, the advent of
big data, and computational methods
e.g.~\cite{Hinton1985,Rumelhart:1986,LeCun1989,LeCun1998,bengio2003neural,srivastava2014dropout,LeCun2015}
have made it possible to apply (deep) NNs to a plethora of regression
problems~\cite{Goodfellow-et-al-2016}. While it is extremely useful to obtain a
prediction function $y = f(\theta; x)$ with parameters $\theta$ in the first
place, one is often interested in properties such as minima or maxima of this
function $f$.

For example, take radiotherapy cancer treatment. In a first step, one would
like to find out how the output $y$, the tumor size, depends on the input $x$,
the kinds and dosages of certain radiotherapies. Once the NN is trained via
fitting the parameter values $\theta$ and can predict the outcome of a certain
treatment to the desired accuracy, one would like to know: what is the ideal
treatment for the patient's cancer? In this case, we want to find the input
vector $x_{\rm min}$ that minimizes the tumor size. Were $f$ a simple analytic
function, one would simply proceed by finding extrema via solving for zeros of
the derivative function w.r.t.~$x$. However, due to the NN's complicated
nonlinear functional structure, generally high dimensionality and potential
recursive input dependencies~\cite{lstm} this easily becomes a highly
complicated system of equations that is extremely difficult to
solve.\footnote{For a simple example, consider a single layered neural network
with activation $a(z) = \tanh(z)$. Then the derivative of $a(x) = \tanh(W \cdot
x + b)$ with weights $W$ and biases $b$ allows us to find the optimal vector
$x_{\rm ext}$ for the network.} The strength of NNs also becomes a weakness in a
sense: while NNs can reveal the elusive black box functions presented by certain
processes, it can only do so in a complicated web of weights and biases that
make an analytical analysis of the NN challenging.

In this paper, we present an alternative solution to this problem via a process
that we term \textit{extremal learning}: Finding the extremizing input $\hat x$
is formulated as the training of an NN itself. The parameters $\theta$ of the NN
are frozen while the input vector $x$ is promoted to a trainable variable
itself. A loss function $\mathcal{L}$ is defined such that its minimization,
e.g.~via gradient descent~\cite{Rumelhart:1986,lecun2012efficient}, is
equivalent to achieving an extremal value for $f$. Hence, custom machine
learning frameworks such as \texttt{TensorFlow}~\cite{abadi2016tensorflow} can
be used to calculate extrema of NNs via this method of extremal learning. For an
implementation of extremal learning in \texttt{TensorFlow}, we provide example
code at: \url{https://github.com/ZakariaPZ/Extremal-Learning}.

The loss function $\mathcal{L}$ may also include further constraints on the
problem at hand. A common constraint in regression problems is that $\hat x$
should not stray too far from the input data set $(x_n)$. Respecting this
constraint avoids extrapolating into a no-data regime where the predictions of
the NN are becoming unreliable. This can be achieved by limiting the
extrapolation of the input vector to within some set number of standard
deviations. For instance, consider fitting a car's fuel expenditure as a
function of its speed $v$ using data from low speeds. The data reflects a
situation in which the rolling resistance force $F_r$ that does not depend on
speed is dominating the fuel expenditure. For higher speeds, the force
dominating the fuel expenditure is the drag force $F_d \propto v^2$. Clearly,
extrapolating to higher speeds with a model trained on low speed data will yield
far too low a fuel expenditure. In a similar spirit, we risk making misleading
predictions if we look for extremizing inputs far from the data set the model was
trained on. Our initial data may not confidently reflect the behaviour of our
system at points lying far from our observations. We will demonstrate how to
include these or other constraints into extremal learning via the definition of
custom loss functions as also discussed
in~\cite{Trinh2017gi,2020arXiv200613554M}.

\subsubsection*{Relationship to GANs}

Extremal learning has some similarities with Generative Adversarial Networks
(GANs)~\cite{Goodfellow2014gi, 8253599, Hong2019gi}.  In a typical example of
extremal learning, we have a trained model for image classification and now want
to find an image that maximizes the output for a given class. A GAN operates
with a similar goal, yet executes it differently. Given a random noise input
$z$, the generator in a GAN is trained to minimize the loss
\begin{equation}
 \mathcal{L} = \log\left\{1-D\left[G(z)\right]\right\}\,,
\end{equation}
where $G(z)$ is the output of a generator network and $D$ is a discriminator
network. The goal of the discriminator is to identify which inputs are real and
which inputs are counterfeits of the generator, while the generator attempts to
fool the discriminator into classifying its output as a true input.  By
minimizing the above loss, the generator should learn a generative model capable
of transforming a random noise sample into an output $G(z)$ which closely
resembles the nature of a true instance $y$. Then, if we passed $G(z)$ into a
discriminator network, it should classify the input similarly to that of the
true instance $y$. Succinctly, a GAN works to minimize the dissimilarity between
an artificially produced data distribution and a true distribution. For extremal
learning, instead of a generator-discriminator pair, we consider a lone
discriminator whose loss function we try to minimize. An important distinction
here is that extremal learning directly changes the input vector $x$, rather
than using other optimization methods such as Markov Chains to create a
generative model.

\subsubsection*{Relationship to adversarial training}

We can also draw parallels to adversarial
training~\cite{10.1145/2046684.2046692, Goodfellow2014gi, 2016arXiv161101236K}.
In image classification, adversarial attacks involve injecting an adversarial
vector $x_{\rm adv}$ into the input vector $x$, where $x_{\rm adv}$ acts as a
perturbation, imperceptible to the human eye yet causing misclassification in a
machine learning model. Adversarial training aims to achieve the generation of
such noise vectors $x_{\rm adv}$, known as adversarial examples. This involves
training an input vector $x$. One technique to generate adversarial examples is
the fast method~\cite{Kurakin2017gi}. 

This method aims to maximize a linearized loss function $\LL$ that is
approximated to first order as
\begin{equation}
	\LL(\tilde{x}; \theta) \approx \LL(x; \theta) +
	(\tilde{x} - x)^T\nabla_x \LL(x;\theta)\,,
\end{equation}
where $\tilde{x}$ is the perturbed vector, i.e.~the adversarial example in
training. The above loss is maximized subject to the following max-norm
constraint on $\tilde{x}$~\cite{2014arXiv1412.6572G}:
\begin{equation}
||\tilde{x} - x||_\infty \leq \epsilon\,,
\end{equation}
i.e.~each pixel can only be changed by a value $\epsilon$. The adversarial
example is then generated by training the input as
\begin{equation}
	\tilde{x} = x + \epsilon\, {\rm sign}[\nabla_x \LL(x; \theta)]\,.
\end{equation}

The extremal learning technique introduced in this paper also seeks to train
inputs, but instead hopes to minimize a nonlinearized loss function via
backpropagation. The input perturbations are not limited by a max-norm
constraint, but the learning rate of the backpropagation.

\subsection*{A road map}

This paper is structured as follows: in Section~\ref{sec:findingextrema}, we
describe the general formalism of extremal learning for regression problems. In
Section~\ref{sec:example}, we present a toy example of this approach by finding
a maximizing input vector using an implementation of extremal learning in
\texttt{TensorFlow}. We conclude in Section~\ref{sec:conclusions}.

\section{Finding extrema}
\label{sec:findingextrema}

We first recap supervised learning in regression problems: given a set of
labeled observations $(x_n)$ and $(y_n)$ with an NN function $f(\theta;x)$ with
parameters $\theta$, for supervised learning, the goal is to find the best-fit
parameters
\begin{equation}
	\hat \theta = \arg\min_{\theta} \left\{ \sum_n \LL_{t}\left[f(\theta;x_n),
	y_n\right]\right\}.
\end{equation}
Here, $\LL_t(y', y)$ is the training loss function that measures the distance
between the predicted output $y'=f(\hat\theta;x)$ is from the true output $y$.
Common choices in regression problems include mean squared error (MSE), mean
squared logarithmic error or mean absolute error~\cite{Goodfellow-et-al-2016}.
Once a model has been trained, i.e.~once we have a set of parameters
$\hat\theta$, predictions can be made. Given an input vector $x$, evaluating the
function $f(\hat\theta;x)$ is called inference - the rules dictating how
$f(\hat\theta;x)$ behaves are now approximately encoded within the NN’s
parameters.

\subsection{Extremal learning}
\label{sec:extremal}

In this paper, we want to find an extremizing input $\hat x$ because we are
interested in maximizing or minimizing the output of $f$ that can represent a
quantity that we would like to optimize as for instance a return of investment,
effectiveness of a treatment, amount spent etc. We formulate this extremization
as yet another optimization problem: given a trained model $f(\hat\theta;x)$, we
want to find the input(s) $\hat x$ that maximize the output of the model, i.e.
\begin{equation}
	\hat x = \begin{cases} \arg\max_{x} f(\hat\theta;x) & \text{for
		maximization}\,,\\
		\arg\min_{x} f(\hat\theta;x) & \text{for minimization}\,.
	\end{cases}
\end{equation}

We call this task extremal training, because we are looking for an input vector
that is extremizing the output. We can implement this optimization as fitting
$f$ via the following steps:
\begin{itemize}
	\item Freeze the parameters $\theta$ of the model to the value $\hat
		\theta$ from the first training iteration, i.e.~make them not
		trainable anymore.
	\item Promote the input vector $x$ to a trainable variable. From the
		fitting procedure point of view, there is no input data anymore
		as $\theta=\hat\theta$ is fixed and $x$ is now effectively a
		parameter. We now have a different NN, albeit with the same
		architecture as the original NN.
	\item Define a loss function $\hat \LL$ such that its minimization is
		equivalent to the optimization problem at hand, i.e.~finding the
		extremizing input $\hat x$. Reasonable, MSE inspired, loss
		functions are
		\begin{equation}
			\hat\LL = \begin{cases} \left[f(\hat\theta;x)^2
			+\kappa\right]^{-1} & \text{for maximization}\,, \\
			f(\hat\theta;x)^2 & \text{for
			minimization}\,,\end{cases}
		\end{equation}
		with constant $\kappa>0$ to avoid the zero divergence.
	\item Minimize the loss $\hat\LL$ via common fitting procedures, such as
		gradient descent and
		backpropagation~\cite{Rumelhart:1986,lecun2012efficient}. The
		gradient descent rule is applied to update the input vector
		$x$ using the partial derivatives of the
		loss with respect to the input $x$~\cite{Yan1988}.
		\begin{equation}
			\label{eq:gradientdescent}
			x\rightarrow x - \alpha\nabla_x \hat\LL\,, 
		\end{equation} 
		where $\alpha$ is the learning rate. One has to provide an
		initial vector $x_{\rm init}$ to start the gradient descent.
		This may be chosen at random or from the training data and may
		be chosen with caution in case of
		non-convexity~\cite{LeCun2015}.
\end{itemize}

\subsection{Constraints via additional loss functions}
\label{sec:addloss}

We may also introduce a variety of $k$ additional loss functions $\LL_i$ to
penalize undesirable inputs $x$ or outputs $y$
\begin{equation}
 \LL = \hat\LL + \sum_{i=1}^{k} \LL_i\,.
\end{equation}
In practice, one may introduce as many $\LL_i$ as desired to constrain the
search for $\hat x$ to a space that is desirable for the problem at hand for
both the inputs and outputs. 

A common constraint is that the global or local extrema of $f$ may not be the
most feasible solution to the optimization problem.  Optimizing an input demands
that we appreciate that there exists some boundary beyond which the input may
become unrealistic. This is the extrapolation problem: we do not want to
extrapolate too far from the original data set $(x_n)$ as the predictions
$f(\hat\theta;x)$ may become arbitrarily unrealistic. Using extremal learning
in conjunction with custom loss functions $\LL_i$ facilitates convergence
towards an optimum within an error and constraint window that is often more
useful in real-life applications than global or local extrema of the function
$f$.

One way to define a loss function that penalizes extrapolation is as follows: if
$x$ deviates more than $c$-times the standard deviation $\sigma$ of the data set
$(x_n)$ from the mean of the data set, those input vectors are penalized:
\begin{equation}
	\label{extrapolationloss}
	\LL_1 =\kappa_1 \sum_{i=0}^{m}\begin{cases} 
	  (x_i - \mu_i + c\sigma_i)^2 & x_i < \mu_i - c\sigma_i\,,  \\
          0 & \mu_i - c\sigma_i\leq x_i \leq \mu_i + c\sigma_i\,, \\
	  (x_i - \mu_i - c\sigma_i)^2 & x_i  > \mu_i + c\sigma_i\,,
       \end{cases}
\end{equation}
where $x_i$ are the individual components of the $m$-dimensional input vector
$x$ and $\kappa_1$ is a normalization constant. To avoid problems during
gradient descent, one may choose continuous functions for the additional loss
functions $\LL_i$ and ensure their relative weighting through normalization
constants $\kappa_i$ is inline with how strict the different constraints should
be enforced relative to each other~\cite{Trinh2017gi,2020arXiv200613554M}.

\begin{figure}[!htb]
\centering
\includegraphics[width=.49\textwidth]{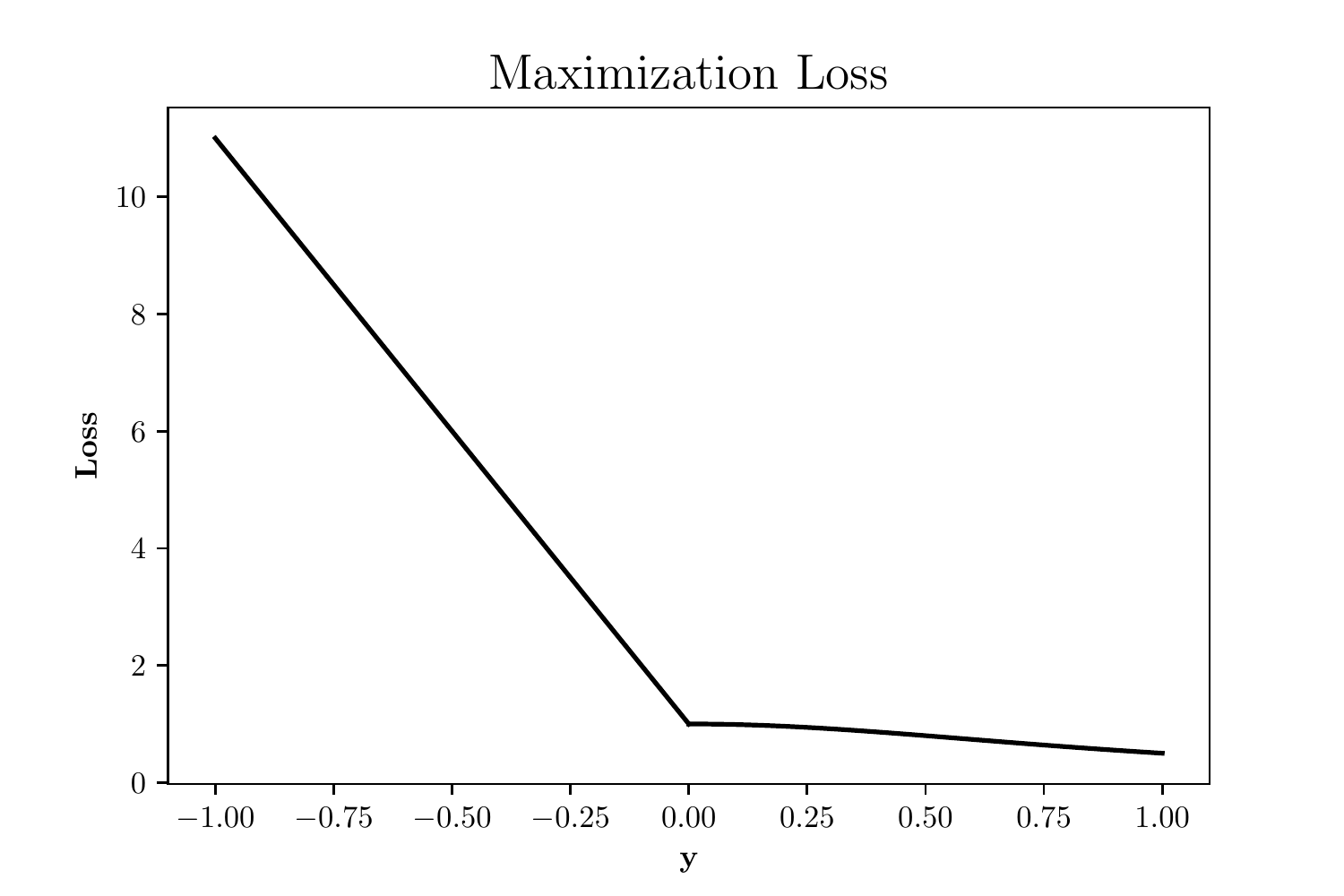}
\includegraphics[width=.49\textwidth]{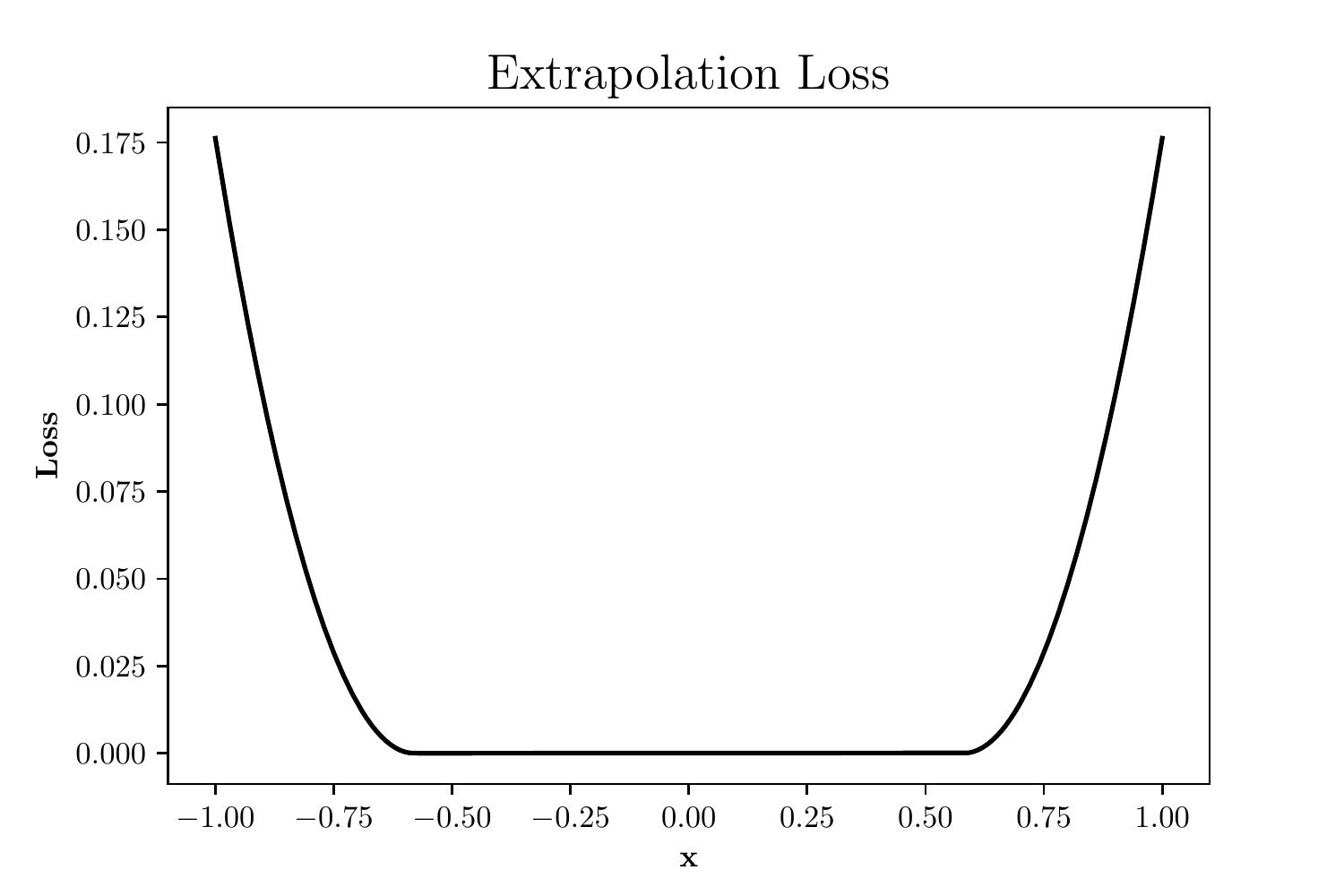}
	\caption{(a) Plot of maximization loss defined
	in~\eqref{maximizationloss} with $\kappa_2 = 10$ and $\hat\kappa = 1$.
	(b) Plot of extrapolation loss defined in~\eqref{extrapolationloss} with
	$\kappa_1 = 1$.}
	\label{fig:losses}
\end{figure}

One may also combine additional constraints with the extremal loss function
$\hat\LL$. For instance, let us consider a situation in which negative outputs
are ill defined as is the case for example if $y$ represents a temperature that
we want to maximize. In
this case, we can define the loss function as
\begin{equation}
 \label{maximizationloss}
 \LL_2 = \begin{cases} 
           -\kappa_2 y+\hat\kappa& y < 0\,, \\
		\left(y^2+\hat\kappa^{-1}\right)^{-1} & y  \geq 0\,,
       \end{cases}
\end{equation}
where the first term ensures that the output is indeed positive. To give this
physical constraint priority over maximization we would choose $\kappa_2 \gg
\hat\kappa$. For a visualization of the maximization and extrapolation loss, see
Figure~\ref{fig:losses}.

If (some of) the components of the input vector are required to be
positive this can be enforced via a loss function similar
to~\eqref{maximizationloss}:
\begin{equation}
 \LL_3 = \kappa_3 \sum_{i=0}^{m} \begin{cases} 
           -x_i & x_i < 0\,, \\
		0 & x_i  \geq 0\,,
       \end{cases}
\end{equation}

\section{Example and case study}
\label{sec:example}

We now consider a simple toy example to demonstrate extremal learning. We first
generate fake data that is used to fit a feed forward NN. After fitting, we apply
the extremal learning approach described above to find a maximizing input. We
provide the source code to this example using \texttt{TensorFlow} at:
\url{https://github.com/ZakariaPZ/Extremal-Learning}.

Consider a simple problem in which $x \in \mathbb{R}^4$ represents a specific
intake of food classes to be optimized with respect to some arbitrary measure of
“goodness of health” $y$. The classes are as follows: 
\begin{itemize}
  \item $x_0$: carbohydrate intake 
  \item $x_1$: protein intake
  \item $x_2$: cake intake
  \item $x_3$: candy intake
\end{itemize}
All inputs are chosen to be in the range $x_i \in [-1,1]$. We want to know how
health depends on the different intakes and, in the next step, determine which
intake composition is the most optimal for health.

\subsection{Data generation}
\label{sec:datageneration}

To generate a data set, we come up with a function that measures "goodness of
health" as a function of the four different inputs in arbitrary units:
\begin{equation}
	\label{gtrue}
	g_{\rm true}(x) = 1 - |x_0| - x_1^2 - x_2 - e^{x_3} + \epsilon\,.
\end{equation}
In moderation, $x_0$ and $x_1$, representing protein and carbs respectively, are
good to intake, but as with anything, excess is deleterious to our health.
$x_2$ and $x_3$ contribute negatively - the more intake of these foods, the
worse your health becomes. The $\epsilon$ term contributes Gaussian noise with
mean $\mu = 0$ and standard deviation $\sigma = 0.05$.  Note that the function
has no biological foundation - it is purely for data generation in this toy
example. 

\begin{figure}[!htb]
\centering
\includegraphics[width=.49\textwidth]{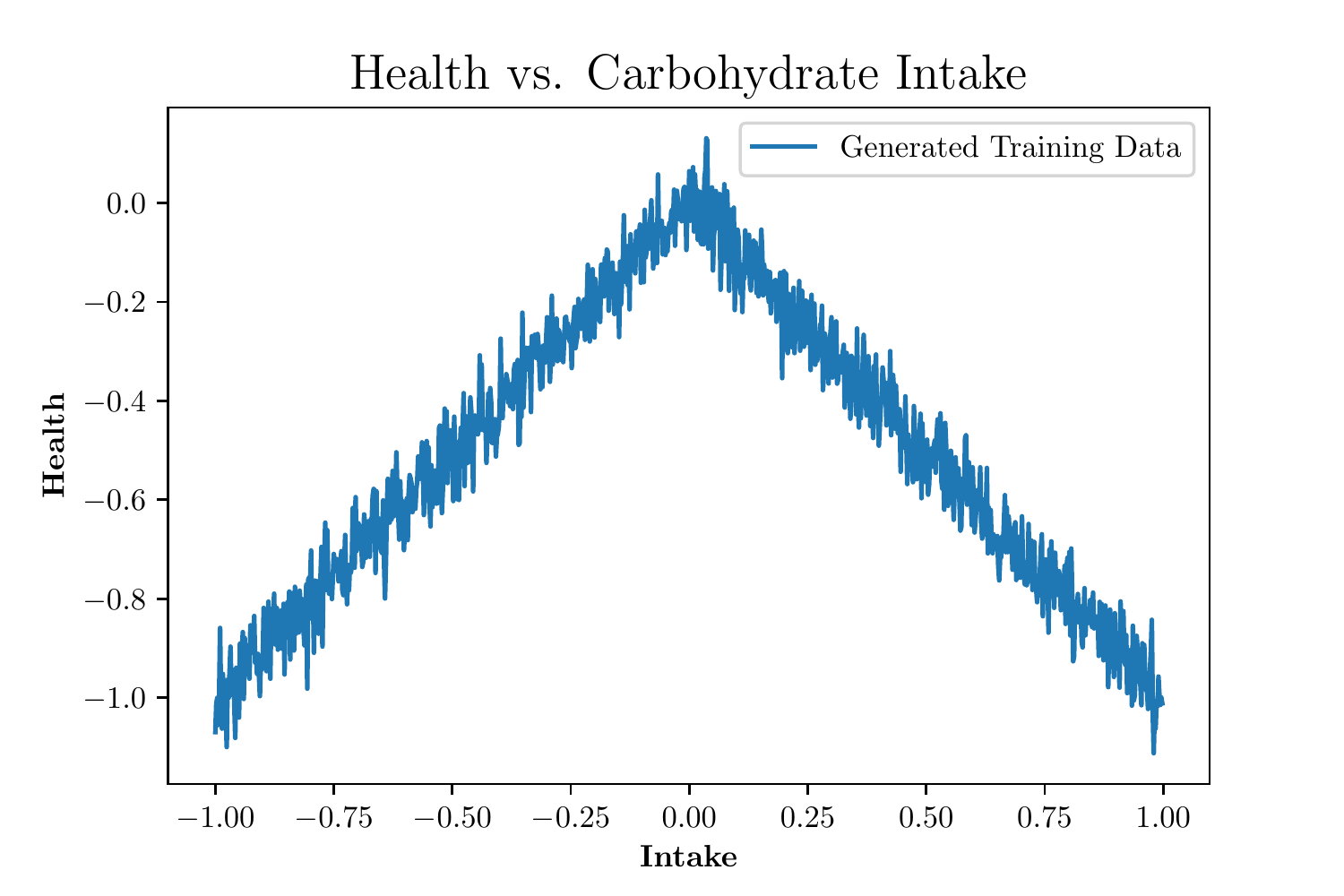}
\includegraphics[width=.49\textwidth]{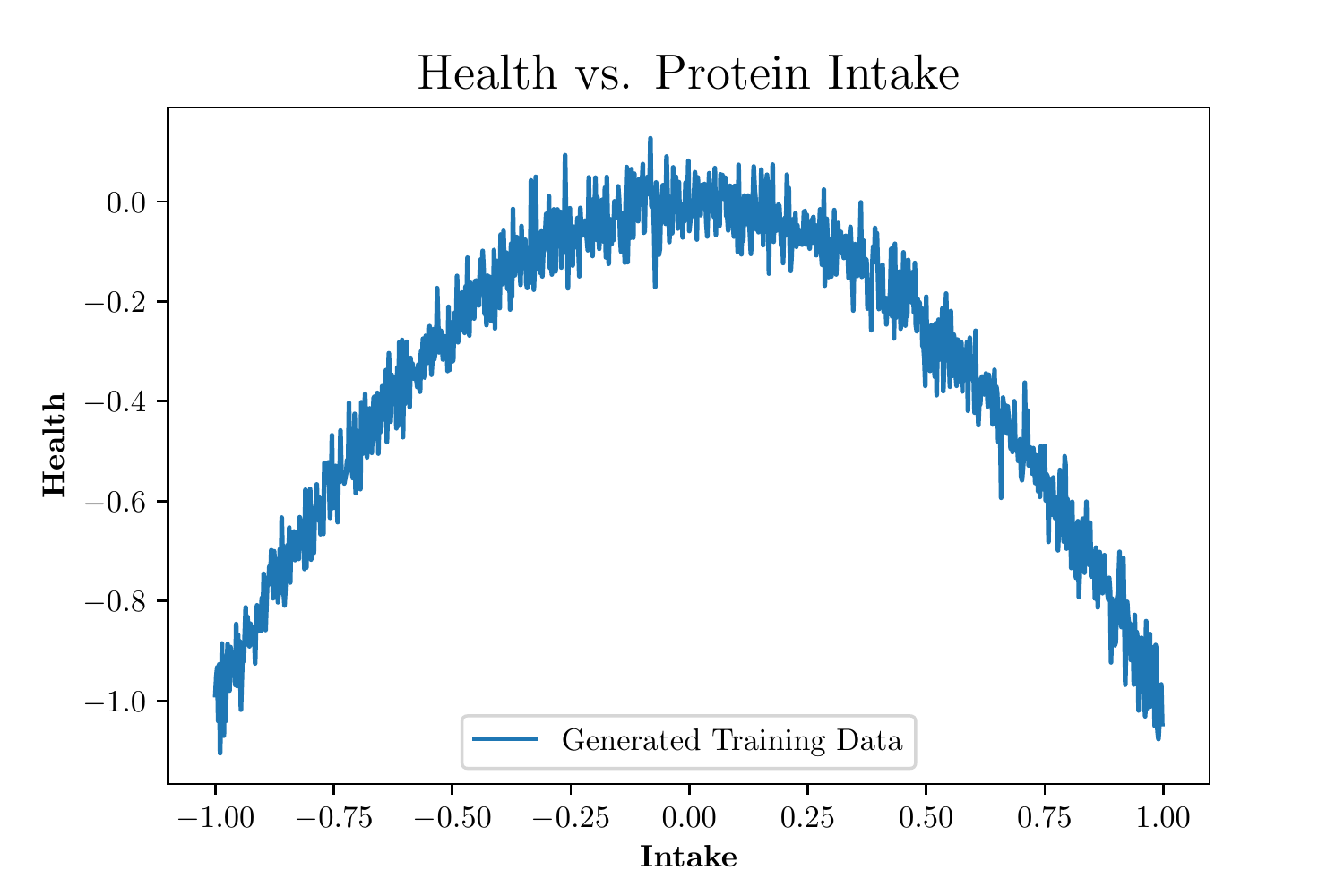}
\includegraphics[width=.49\textwidth]{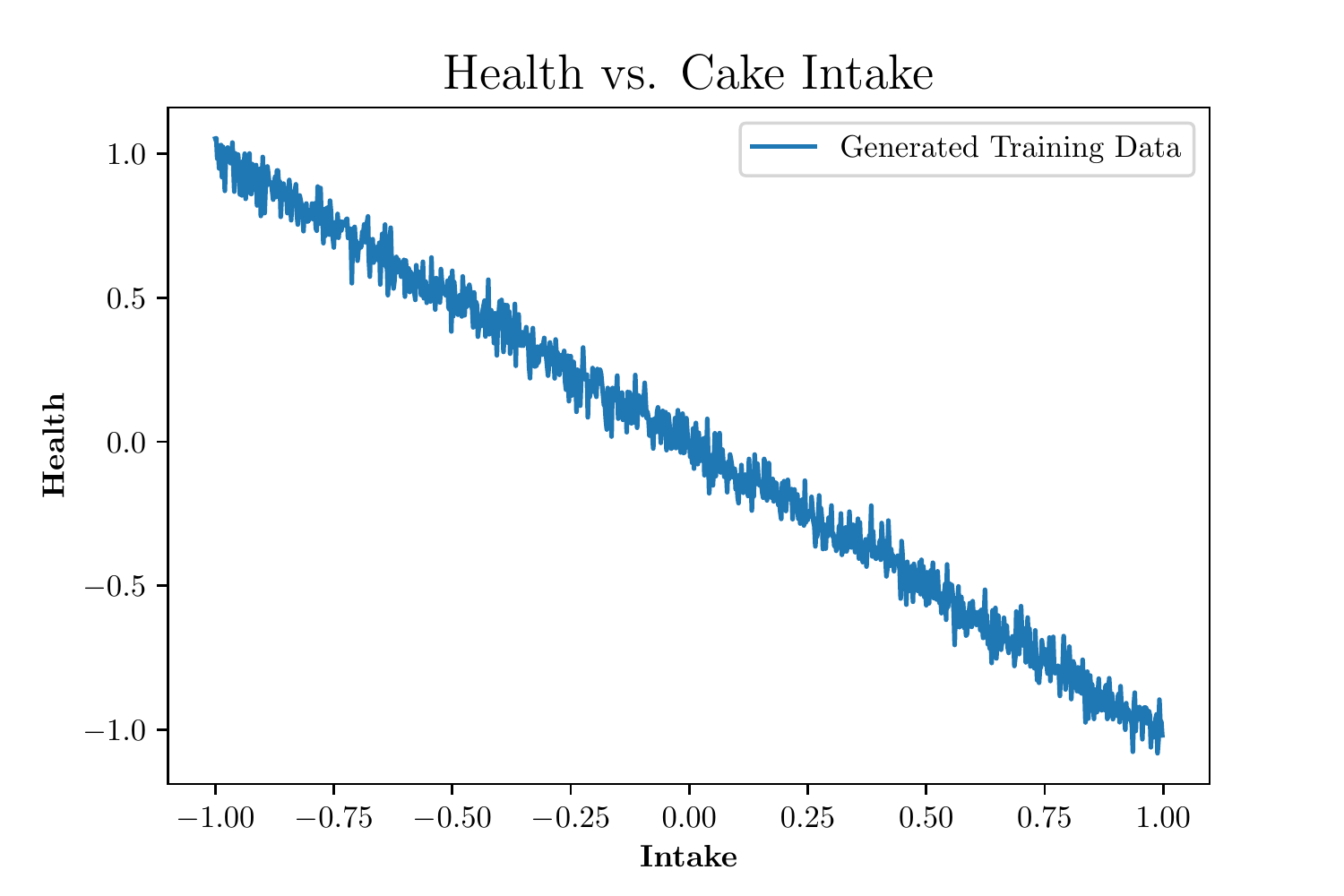}
\includegraphics[width=.49\textwidth]{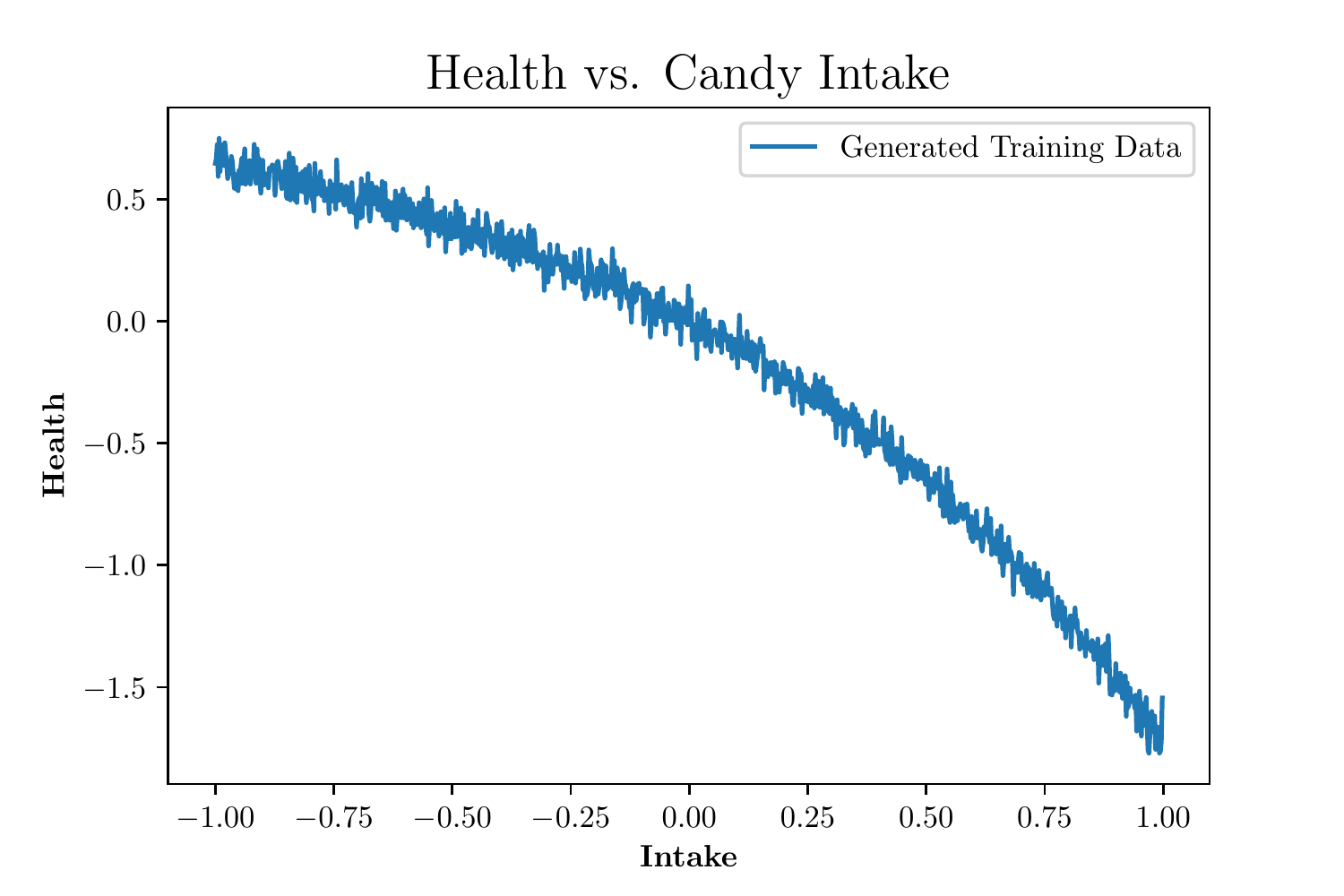}
	\caption{\label{fig:datafeature} Each input $x_i$ is plotted against the
	output health $y$. We want to maximize the output with respect to some
	constraints.  Accordingly, we see that both $x_0$ and $x_1$ maximize $y$
	at $x_0=x_1=0$, while $x_2$ and $x_3$ formally maximize $g_{\rm true}$
	at $x_2=x_3=-\infty$ (though of course, our extrapolation constraint
	impose limits on the domain within which we will search for a
	solution).}
\end{figure}

The data set $(x_n)$ and $(y_n)$ with $y_n=g_{\rm true}(x_n)$ is created by
sampling~\eqref{gtrue} over $n=1000$ inputs sampled from a uniform distribution
where $-1 \leq x_{n,i} \leq 1$. We plot the data set as health vs the various
inputs in each dimension in Figure~\ref{fig:datafeature}.

\subsection{Finding a maximizing input}

\begin{figure}[!htb]
\centering
\includegraphics[width=.49\textwidth]{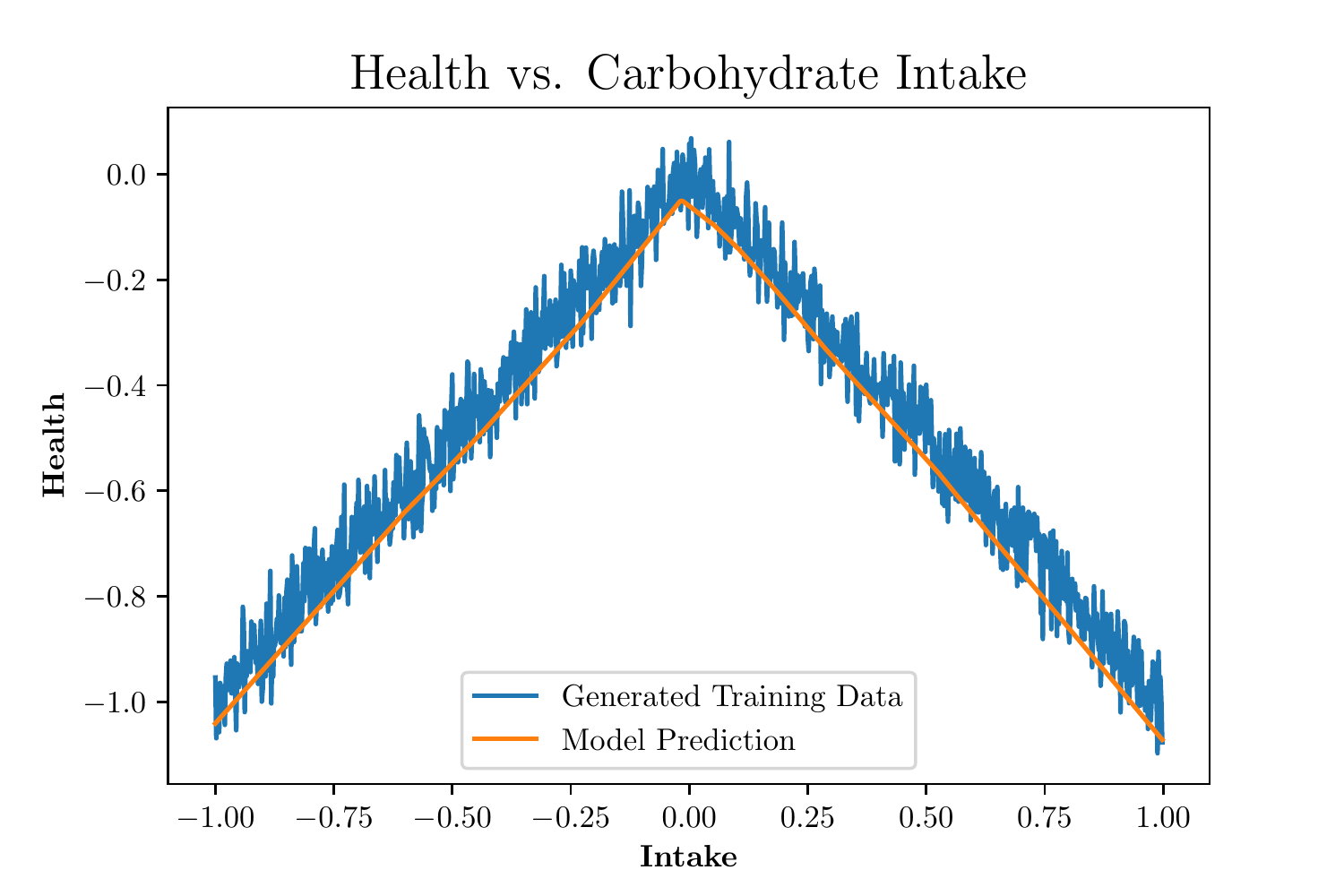}
\includegraphics[width=.49\textwidth]{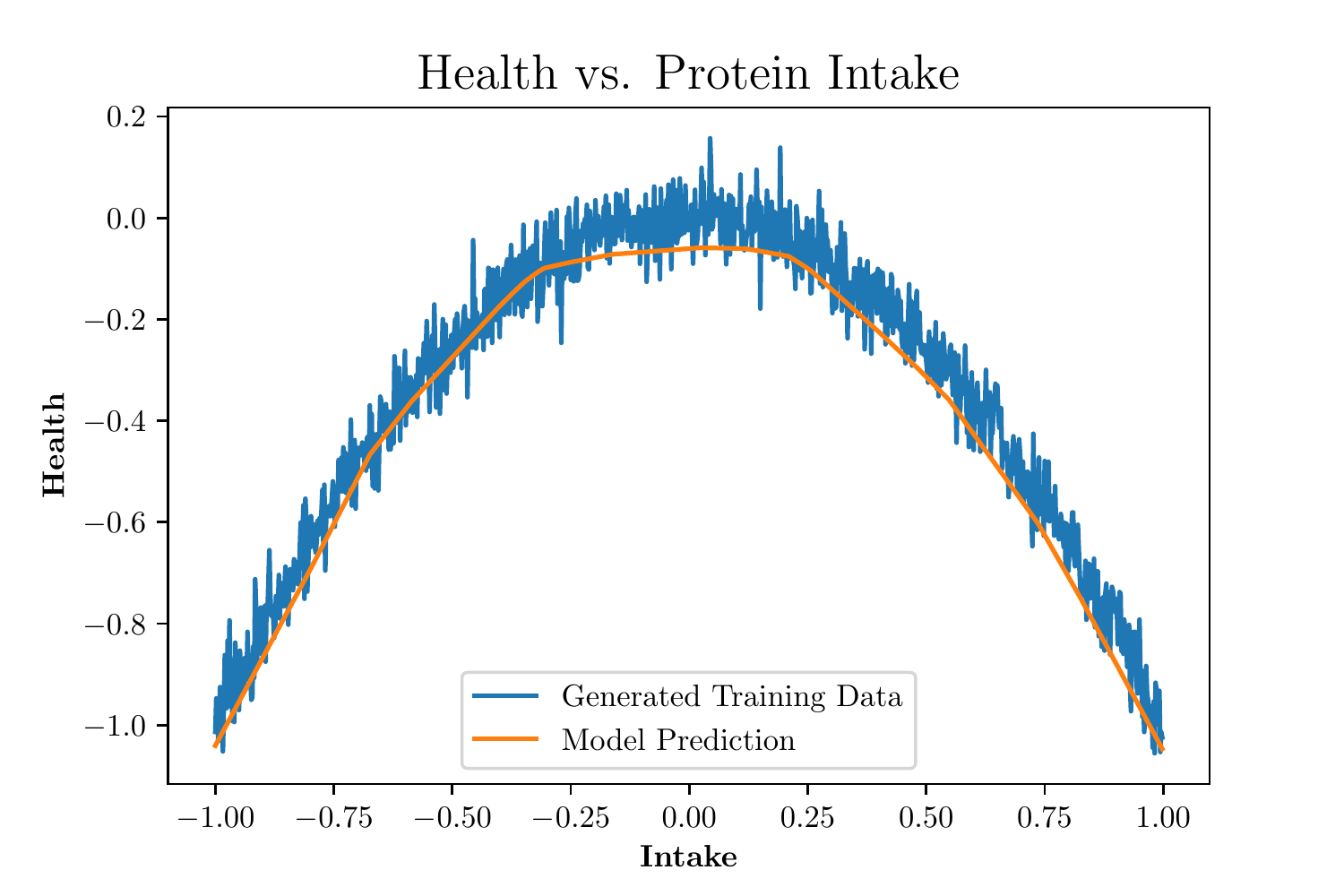}
\includegraphics[width=.49\textwidth]{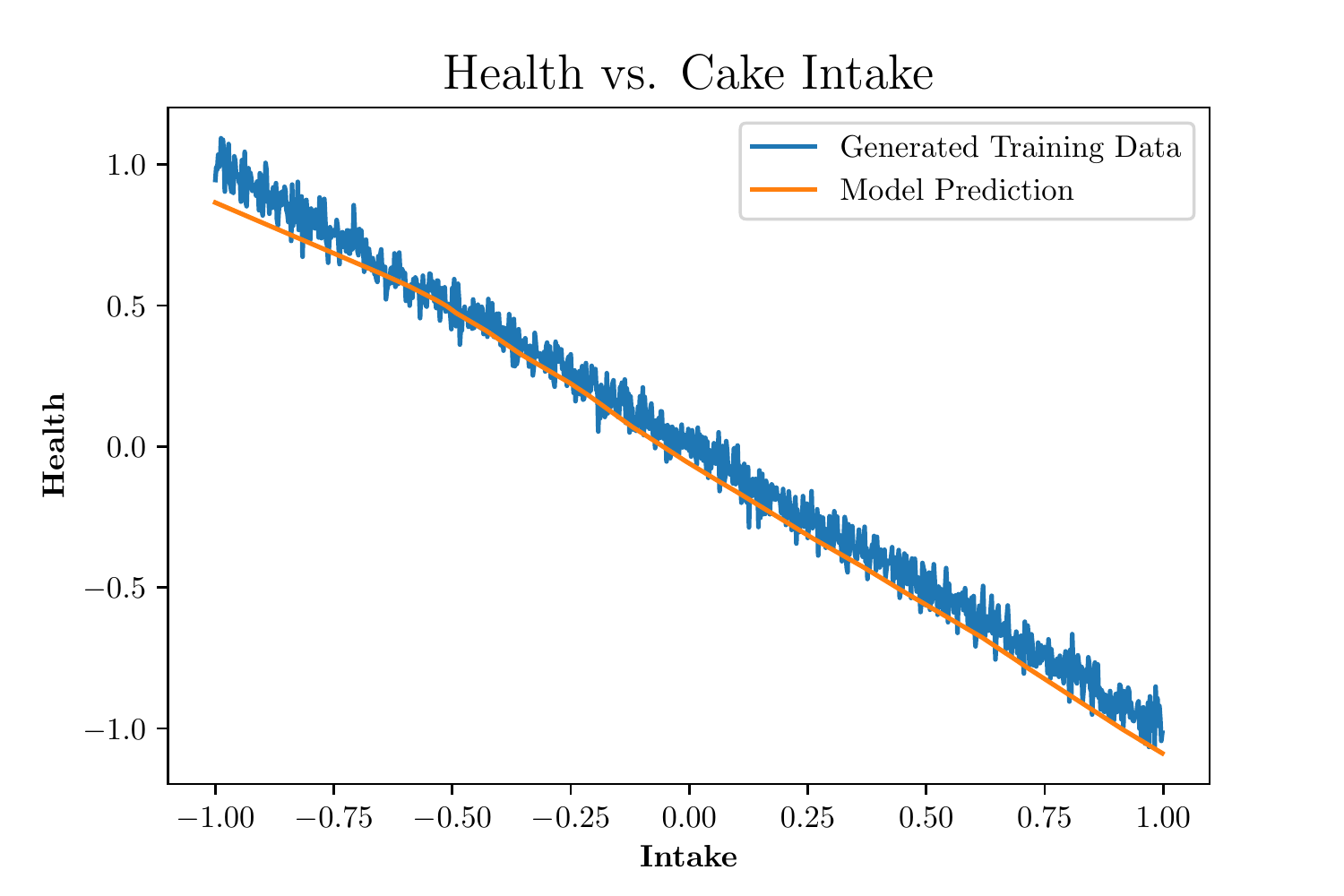}
\includegraphics[width=.49\textwidth]{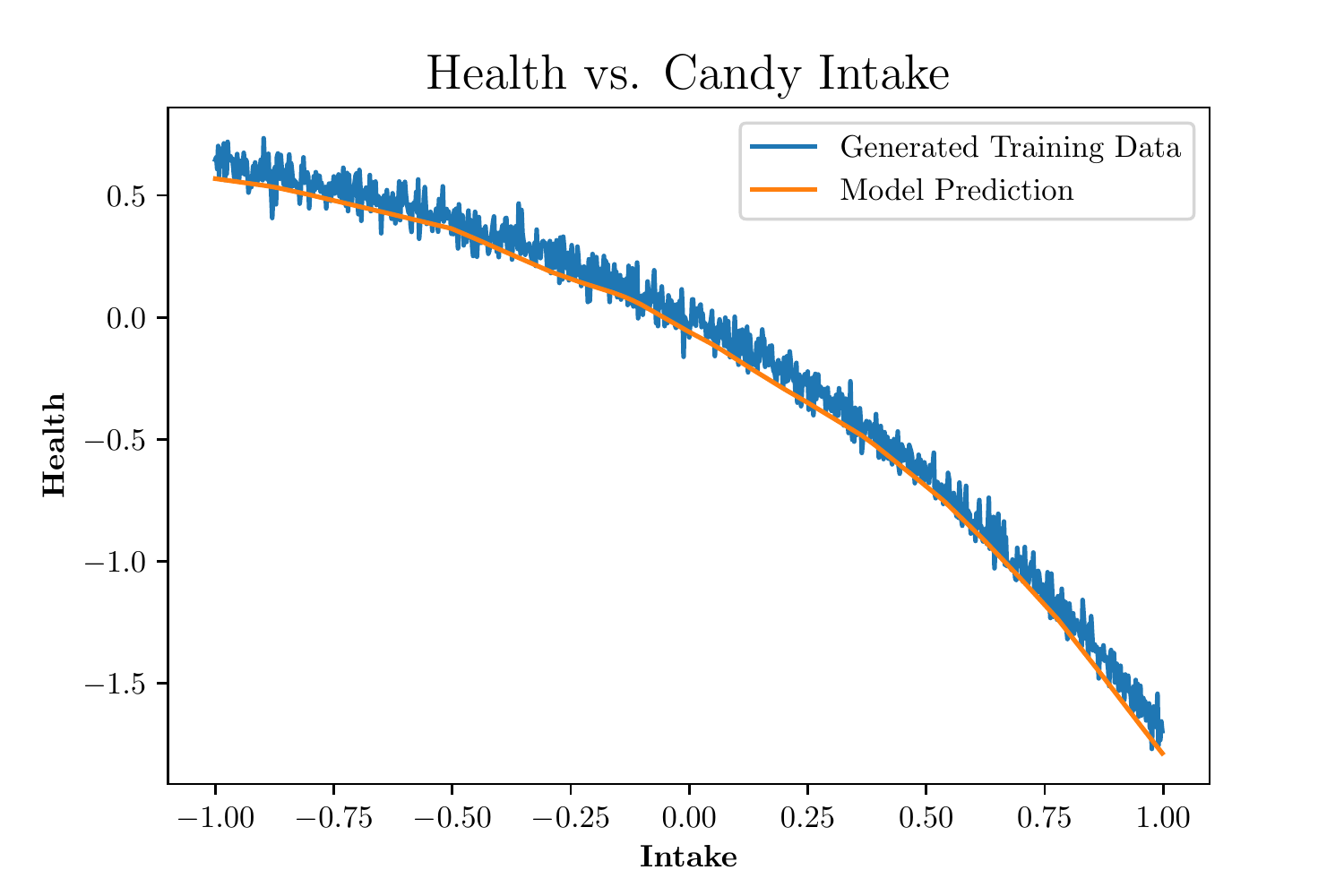}
\caption{\label{fig:datafit} The trained NN is superimposed onto the training
  data for each intake $x_i$. The NN describes the underlying functional
  dependence $g_{\rm true}$ of the data to sufficient accuracy.}
\end{figure}

Ideally, we would like the network to learn the underlying functional
dependence~\eqref{gtrue}. To this extend, we define a feed forward NN
$f(\theta;x)$ that is trained via gradient descent on the data $(x_n)$, $(y_n)$.
Subsequently, we have a set of parameters $\hat\theta$ containing the weights
and biases of the NN which are conducive to a good approximation of $g_{\rm
true}$. We show the fitted NN in Figure~\ref{fig:datafit}.

Following~\ref{sec:extremal}, we can now apply extremal learning. First we
freeze the parameters $\theta$ to $\hat\theta$, i.e.~they are not trainable
parameters anymore. Secondly, we promote the input $x$ to trainable parameters.
We now have a second NN that has identical architecture to the original NN with
the only difference being what is considered a parameter and what is considered
an input. The extremal learning NN has no input, just trainable parameters $x$
that were the input of the previous NN.

When training the extremal learning NN, one has to provide a starting value
$x_{\rm init}$. We randomly initialize the gradient descent with an input vector
$x_{\rm init}$ from a normal distribution such that $x_{{\rm init}, i} \in
[-1,1]$.

Next, we set constraints using additional loss functions defined
in Section~\ref{sec:addloss} tailored to this specific
problem. We use the extrapolation loss function~\eqref{extrapolationloss} and
penalize in the search for the optimal input if it is more than two standard
deviations away from the mean. For the maximizing loss we
use~\eqref{maximizationloss} with $\kappa_2=10$ and $\hat\kappa=1$. Hence, the
total loss is
\begin{equation}
	\LL = \LL_1 + \LL_2\,,
\end{equation}
with
\begin{equation}
\LL_1 =\frac12\sum_{i=0}^{3}\begin{cases} 
	  (x_i - \mu_i + 2\sigma_i)^2 & x_i < \mu_i - 2\sigma_i\,,  \\
          0 & \mu_i - 2\sigma_i\leq x_i \leq \mu_i + 2\sigma_i\,, \\
	  (x_i - \mu_i - 2\sigma_i)^2 & x_i  > \mu_i + 2\sigma_i\,,
       \end{cases}
\end{equation}
 and
\begin{equation}
     \LL_2 = \begin{cases} 
           -10\, y+1& y < 0\,, \\
		\left(y^2+1\right)^{-1} & y  \geq 0\,,
       \end{cases}
\end{equation}

The mean and standard deviation of the data set generated in
Section~\ref{sec:datageneration} numerically evaluate to~\footnote{In the limit
$n\rightarrow \infty$, we expect $\mu = [0, 0, 0, 0]$ as the data is sampled
from a uniform distribution with mean zero. However, we see some noise remnants
due to finite $n=1000$. The standard deviation of a uniform distribution sampled
over the interval $a \leq x \leq b$ is approximately
$\sigma=\frac{(b-a)}{\sqrt{12}}[1, 1, 1, 1]$, which evaluates to
$\sigma\simeq[0.577, 0.577, 0.577, 0.577]$ in this case.}

\begin{align}
    \begin{aligned}
	    \mu &\simeq [0.007,  -0.028, 0.005,  0.006]\,,\\
	    \sigma &\simeq [0.555, 0.577, 0.577 , 0.567]\,.
    \end{aligned}
\end{align}

From here, we perform gradient descent to train the extremal learning NN via
subsequently applying~\eqref{eq:gradientdescent} until we converge on a
maximizing input $\hat x$. We expect that the extremal NN's output should
approach a maximum which is not the analytical maximum since we are using a numerical
technique with additional constraints arising from the loss functions. However,
considering the form of $g_{\rm true}$, there is no true maximum as decreasing
$x_1$ and $x_2$ towards $-\infty$ will continuously increase the value of $g_{\rm
true}(x)$. We should instead see that the new output of the input-optimized
model converges towards a value limited by the extrapolation loss. Indeed, we
numerically find the maximizing input to be
\begin{equation}
 \hat x \simeq [-0.167,  -0.0861, -1.193, -1.153 ]\,,
\end{equation}
which is reasonably close to the true optimal value under the limits of the
extrapolation loss we have chosen:
\begin{equation}
	\hat x_{\rm true} = [0, 0, -2\sigma_2, -2\sigma_3] \simeq [0, 0, -1.154,
	-1.134]\,.
\end{equation}

Similarly, the optimal output value calculated via extremal learning $\hat
y\simeq1.702$
is reasonably close to the true optimal value of the underlying function in this
optimization problem $\hat y_{\rm true} = 1 + 2\sigma_2 - e^{-2\sigma_3} \simeq
1.832$

\section{Conclusions}
\label{sec:conclusions}

In this paper, we have introduced extremal learning which allows to calculate
extremizing inputs of a trained NN in regression problems. This is a pressing
issue in many applications of modern machine learning, where one is not just
interested in inference, i.e.~making predictions from a trained NN, but also
finding the input vector that extremizes a certain output. Examples are finding
the optimal cancer treatment after an NN has been trained that represents how
the patient's tumor size responds to different cancer treatments, or the toy
example discussed in Section~\ref{sec:example}: what is the healthiest diet for
an individual based on an NN that evaluates health as a function of the intake
of various food classes. There are many other real-world examples where solving
extremization questions of this kind are very valuable.  Analytically solving
for extrema is generally not a feasible option as one is dealing with a coupled
system of nonlinear equations of high dimensionality, with potentially recursive
structure~\cite{lstm}, due to the generally complicated underlying functional
dependence defining NNs. 

Extremal learning relies on the same NN infrastructure created to train the NN
network in the first place to perform the optimization task of extremization.
This way we can take advantage of how machine learning overcomes the curse of
dimensionality which also plagues the task of extremization. The basic
components of extremal learning are freezing the parameters such as weights and
biases of the original NN and promoting the input vector $x$ to a trainable
variable. Via choosing an appropriate loss function, common machine learning
optimization techniques such as gradient descent and
backpropagation~\cite{Rumelhart:1986,lecun2012efficient} can then be used
to compute the extremizing input. Common machine learning frameworks such as
\texttt{TensorFlow} have sufficient flexibility to perform this task with
minimal configuration effort,
see~\url{https://github.com/ZakariaPZ/Extremal-Learning}. While there are
certain parallels with GANs~\cite{Goodfellow2014gi, 8253599, Hong2019gi} and
adversarial training~\cite{10.1145/2046684.2046692, Goodfellow2014gi,
2016arXiv161101236K}, extremal learning presents an efficient way to find
extrema of NNs in regression problems.

We also demonstrate how to incorporate further constraints on the input and/or
output vector via additional loss functions, see Section~\ref{sec:addloss}. A
common feature of many extremization tasks is to constrain the input vector not
to measure too far a distance from the data set the NN was originally trained on.
The distance from the original data set can be penalized via such an additional
loss function and, in practice, prevent the NN to look for an extremizing input
where the predictions may become unrealistic. We further demonstrate that it is
straightforward to further constrain the input or output to one's liking via
additional loss functions.

In the future, we would like to study the numerical performance of
extremal learning in a variety of regression examples. An interesting question
remains what extremal learning can add in the context of convolutional neural
networks, in particular if it can be combined with GANs and/or adversarial
training to solve input optimization problems.

\acknowledgments

We thank Sergei Bobrovskyi, Patrick Gonz\'alez, and Sebastian Wetzel
and in particular Malte Nuhn for helpful discussions.

\bibliography{Extrema-ML.bib}
\bibliographystyle{JHEP.bst}

\end{document}